\documentclass[10pt, a4paper]{article}
\usepackage{lrec2022} 
\usepackage{multibib}
\newcites{languageresource}{Language Resources}
\usepackage{graphicx}
\usepackage{tabularx}
\usepackage{soul}
\usepackage{multirow}
\usepackage{float}
\usepackage{titlesec}
\titleformat{\section}{\normalfont\large\bfseries\center}{\thesection.}{1em}{}
\titleformat{\subsection}{\normalfont\SmallTitleFont\bfseries\raggedright}{\thesubsection.}{1em}{}
\titleformat{\subsubsection}{\normalfont\normalsize\bfseries\raggedright}{\thesubsubsection.}{1em}{}
\renewcommand\thesection{\arabic{section}}
\renewcommand\thesubsection{\thesection.\arabic{subsection}}
\renewcommand\thesubsubsection{\thesubsection.\arabic{subsubsection}}

\usepackage{epstopdf}
\usepackage[utf8]{inputenc}

\usepackage{hyperref}
\usepackage{xstring}

\usepackage{color}

\title{MuLVE, A Multi-Language Vocabulary Evaluation Data Set}

\name{Anik Jacobsen\textsuperscript{1}, Salar Mohtaj\textsuperscript{1,2}, Sebastian M\"oller\textsuperscript{1,2}} 

\address{\textsuperscript{1} Technische Universit\"at Berlin, Berlin, Germany \\
        \textsuperscript{2} German Research Centre for Artificial Intelligence (DFKI), Projektb\"uro Berlin, Germany \\
         a.jacobsen@campus.tu-berlin.de, \{salar.mohtaj, sebastian.moeller\}@tu-berlin.de\\}

\abstract{
Vocabulary learning is vital to foreign language learning. Correct and adequate feedback is essential to successful and satisfying vocabulary training. However, many vocabulary and language evaluation systems perform on simple rules and do not account for real-life user learning data. This work introduces Multi-Language Vocabulary Evaluation Data Set (MuLVE), a data set consisting of vocabulary cards and real-life user answers, labeled indicating whether the user answer is correct or incorrect. The data source is user learning data from the Phase6 vocabulary trainer. The data set contains vocabulary questions in German and English, Spanish, and French as target language and is available in four different variations regarding pre-processing and deduplication. We experiment to fine-tune pre-trained BERT language models on the downstream task of vocabulary evaluation with the proposed MuLVE data set. The results provide outstanding results of $> 95.5$ accuracy and F2-score. The data set is available on the European Language Grid.
 \\ \newline  }

\begin{document}

\maketitleabstract

\section{Introduction}
Vocabulary learning is an essential part of foreign language learning. Building an extensive vocabulary is necessary to master a foreign language and communicate successfully~\cite{alqahtani2015importance}. To achieve long-term memory of words and their meaning, repetition and appropriate feedback are crucial~\cite{metcalfe2007principles}. We introduce a Multi-Language Vocabulary Evaluation Data Set (MuLVE), including real-life user vocabulary learning data, aiming to improve vocabulary evaluation.  

phase-6 GmbH\footnote{\url{https://www.phase-6.de/}} (hereinafter referred to as Phase6) offers a digital education tool in the domain of language learning. Their service is a vocabulary trainer, available for various digital devices, optimizing vocabulary training for long-term memory. Students can study vocabulary independently and aligned with the content of their school lessons. Phase6 operates in the German market and focuses on pupils.

The area of vocabulary evaluation and training has not been addressed in Natural Language Processing (NLP). Most existing language learning systems operate on simple rules that compare the user's answer to an existing answer, neglecting potential correct answers such as synonyms or various ways of formatting. These inflexible systems lead to user frustration and a limited learning experience, resulting in users losing interest in language learning completely. 

We aim to establish a robust and significant multilingual data set for vocabulary evaluation from available user learning data provided by Phase6 to allow for the development and training of more flexible systems for the task of vocabulary evaluation. Our contributions are as follows:

\begin{itemize}
    \item Multi-Language Vocabulary Evaluation Data Set (MuLVE): a data set containing different variations of vocabulary cards and real-life user answers with a binary label indicating whether the answer is correct or not.
    \item A first experiment and validation of a transformer model trained and tested on the available data set variations.
    \item We make the data set variations available to the research community\footnote{\url{https://live.european-language-grid.eu/catalogue/corpus/9487}}. It can, for example, be used to train and evaluate vocabulary and language evaluation systems.
\end{itemize}

To the best of our knowledge, it is the first data set focusing on vocabulary evaluation.

Next, we discuss related work and relevant corpora in the related domain of paraphrase detection (Section \ref{sec:related_work}). Section \ref{sec:approach} describes the creation of the MuLVE data set, focusing on the retrieval of the data and the annotation process. In Section \ref{sec:format}, we present the format of the data set and how to access it. Section \ref{sec:analysis} analyses the data set distribution. We experiment with the data set and validate the results in Section \ref{sec:exp}. Finally, we conclude the paper and provide future work.

\section{Related Work} \label{sec:related_work}

The research problem is closely related to tasks in the area of NLP. Semantic similarity, usually of sentences or documents, is discussed within paraphrase detection. Paraphrases are semantically identical sentences that convey the same meaning but use different wording. The research in this field goes beyond the sentence level and further considers documents composed of multiple sentences. The task of paraphrase detection has many applications, such as plagiarism detection~\cite{wahle2021plagiarism}, Q\&A systems~\cite{bogdanova2015detecting}, and text summarization~\cite{agarwal2018textmessages}.

\begin{figure*}[t]
\centering
\includegraphics[width=0.95\textwidth]{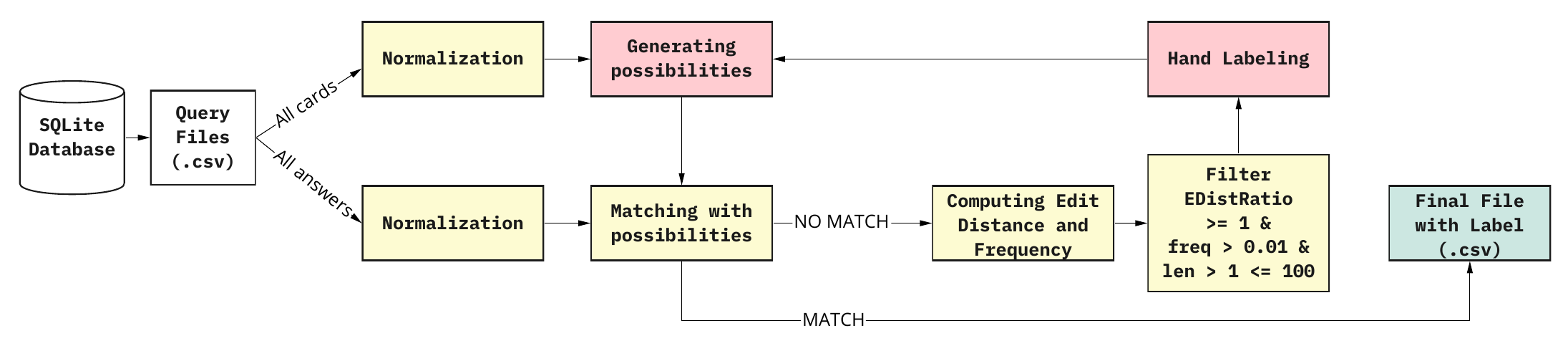}
\caption{Relabeling process.}
\label{fig:relabeling_process}
\end{figure*}

The benchmark corpus in the field of paraphrase detection is the Microsoft Research Paraphrase Corpus (MRPC)~\cite{dolan2005microsoftcorpus}. It consists of 5,801 sentence pairs, where each pair is binary labeled, indicating whether it is a paraphrase or not by a human annotator. The sentences have been collected from newswire articles over two years. The corpus linked with the task of paraphrase detection is included in the GLUE benchmark~\cite{wang2018GLUE}, a collection of nine natural language understanding tasks.

Another relevant corpus, including cross-lingual sentence pairs, is PAWS-X~\cite{yang2019paws}. PAWS-X is derived from PAWS (Paraphrase Adversaries from Word Scrambling~\cite{zhang2019paws}), containing challenging English sentence pairs from Wikipedia and Quora. The noisy paraphrase detection labeled sentence pairs highlight the importance of modeling structure, context, and word order information in the domain of paraphrase identification. PAWS-X contains 23,659 human translated and 296,406 machine-translated PAWS evaluation pairs in six typologically distinct languages: French, Spanish, German, Chinese, Japanese, and Korean. It contains only examples from PAWS-Wiki. In their paper, Yang et al. show the effectiveness of deep, multilingual pre-training on PAWS-X.

Xu et al.~\newcite{xu2015twitterdata} propose a data set consisting of short and noisy texts retrieved from Twitter. It contains 17,790 sentence pairs in the training and development set and 972 sentence pairs in the test set from 500+ trending topics on Twitter (the collection period was between April 24th and May 3rd, 2013). The sentence pairs have a label indicating whether they are paraphrases, not paraphrases, or debatable.

Quora Question Pairs (QQP)\footnote{\url{https://www.kaggle.com/c/quora-question-pairs}} is another data set used to train and evaluate paraphrase detection approaches. It consists of over 400,000 question pairs. Each question pair is annotated with a binary value indicating whether the two questions are paraphrases of each other.

In contrast to the aforementioned data sets, MuLVE focuses on the new task of vocabulary evaluation and contains vocabulary cards and their respective user answers. MuLVE is constructed from real user learning data and aims to improve language evaluation systems.

\section{Approach} \label{sec:approach}
This section explains in detail the creation of the MuLVE data set. Section~\ref{sec:data_ret} provides information on the retrieval of the data set using user learning data from Phase6. Section~\ref{sec:data_anno} describes the data annotation process.

\subsection{Data Retrieval} \label{sec:data_ret}
The source of the data is the Phase6 user input database. Phase6 has collected anonymous user inputs, resulting in more than 450 M available data points. Currently, Phase6 offers support for more than 20 languages. The company focuses on the German-speaking market; thus, most vocabulary questions are in German. The data set will focus on the three most popular target languages: English, French, and Spanish.

\begin{figure}[H]
\centering
\includegraphics[width=0.48\textwidth]{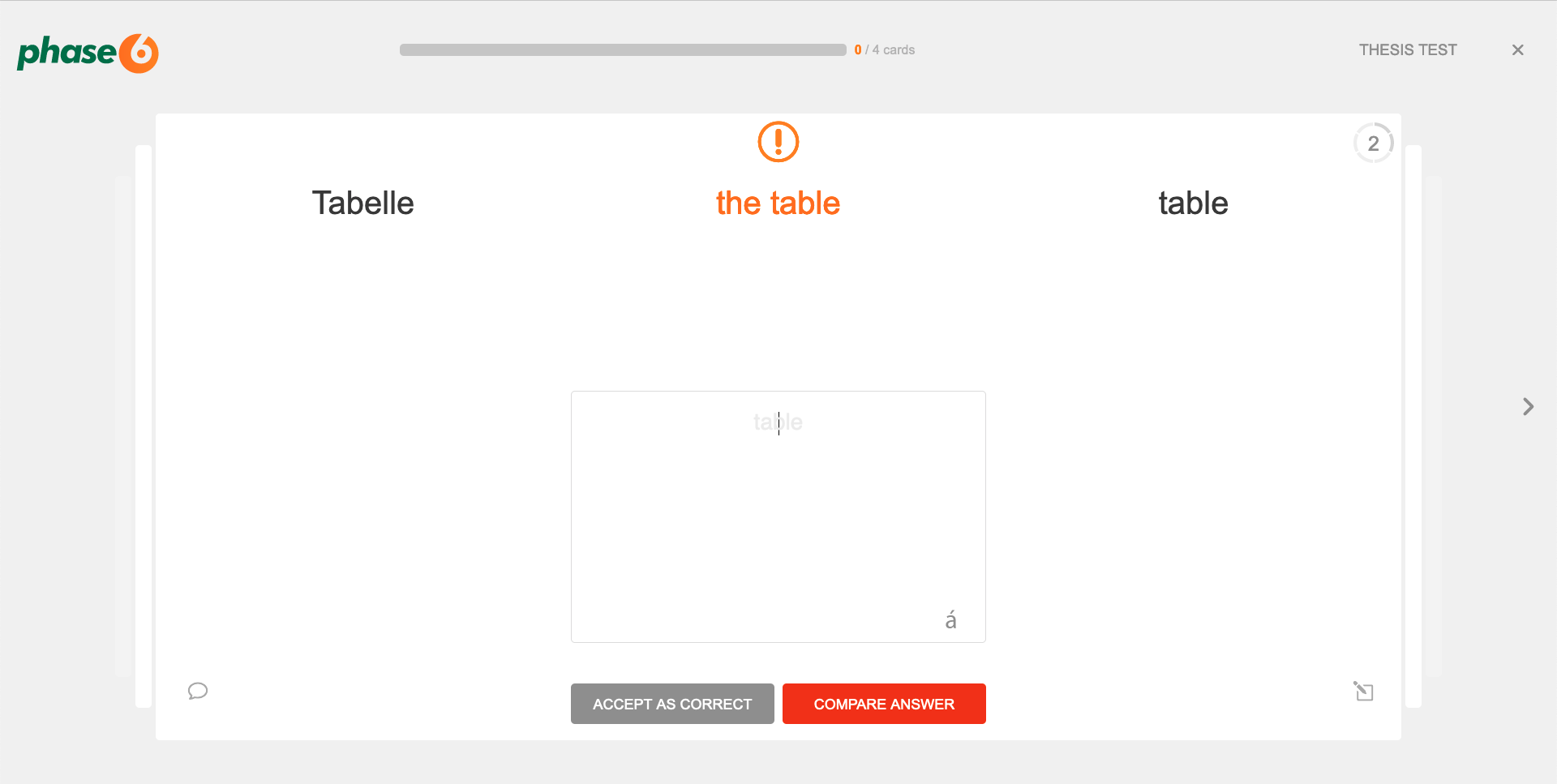}
\caption{Screenshot of the vocabulary learning environment. If an answer is flagged as incorrect, the user can select \textbf{accept as correct}.}
\label{fig:screenshot}
\end{figure}

When a user's answer is flagged as incorrect by the Phase6 system, the user has the option to select \textbf{``accept as correct''} in which case the vocabulary is marked as correct. These data points will be referred to as \textbf{I was right (IWR)}. A screenshot of the user interface can be seen in Figure \ref{fig:screenshot}.

All user answers for the top 1,000 most learned vocabulary cards are included to build a representative and not too extensive data set. Also, the user answers are limited to the IWR and Wrong user answers, as these include misclassifications which we aim to include in the data set. 

\subsection{Data Annotation} \label{sec:data_anno}
The initial approach used the IWR and Wrong classes as labels for the data points. During data exploration, we discovered, however, that user behavior is quite variant; entailing that some users accept the system's decision, while others use the option to select ``Accept as correct'' (even when their answer might certainly be incorrect). Thus, user answers containing the exact same text are present in both the IWR and Wrong class; this results in a low data quality. 

A semi-automatized process for relabeling is shown in Figure~\ref{fig:relabeling_process}. It focuses on generating possibilities of additional correct answers, combined with an extra loop to identify synonyms and other correct solutions which are not yet included. The user answers are then compared with these lists of possible correct answers and labeled \textbf{True} if there is a match and \textbf{False} if there is no match. The individual steps will be discussed in detail.

\subsubsection{Normalization} \label{sec:norm}
First, the data is normalized to make the format as consistent as possible and then compare the user answers with the possibilities. The text is converted to lower case, and everything in parentheses in the answer is ignored, as it mainly contains irrelevant additional information. Punctuation is removed because it is not essential on vocabulary level. Further, spacing is adapted to single spacing, and line breaks are removed. Since children primarily use the vocabulary trainer, wrong versions of the apostrophe character (``~'~'') are replaced; many children are not yet used to the spelling of foreign languages, and apostrophes are rarely used in German. Thus, accents and other characters that appear similar to an apostrophe are corrected.

\subsubsection{Generating Possibilities}
The generation of possibilities encompasses two different aspects. Firstly, it aims to include additional semantically correct solutions such as synonyms. Moreover, it aims to include additional differently formatted correct solutions.

Synonyms are words that have the same semantic meaning and, for example, in English, often occur when comparing British and American English. Another example is the use of different words depending on the desired formality. 

The formatting is not unique for all vocabulary cards. There are several such formatting cases for each language, and we explain a few as examples. Despite these differences in formatting, the system should accept a correct user answer, as we aim for the user to learn the vocabulary and not its format. 

An example in the English language is the use of \textit{``to''} in the case of the \textit{to-infinitive}. There are several ways to answer an English verb vocabulary question correctly. 

In Romance languages, in the case of the available languages French and Spanish, there often exist different versions of nouns and adjectives for the male and female forms. In this case, the male and female, as well as any combination of the two, are correct.

Another common formatting is the use of indefinite pronouns, like: (something = \textit{sth}, somebody = \textit{sb}). If the vocabulary card asks for the answer \textit{``to find sth/sb''}, the indefinite pronouns are not vital for the meaning of the verb when translating it from German. Thus, correct answers are any combination of these pronouns.

These format variations can be generated automatically by defining rules for the possible formatting options. The result is a list of possibilities for each vocabulary card, which can be compared with the user's answers to find possible matches. Examples can be found in Table~\ref{tab:possibility_examples}.

\begin{table}[]
\centering
\resizebox{0.48\textwidth}{!}{%
\begin{tabular}{|l|p{.27\textwidth}|}
\hline
\textbf{answer}         & \textbf{answer possibilities}                                        \\ \hline
till twelve o'clock     & till twelve o'clock, until twelve o’clock                            \\ \hline
(to) ask                & to ask, ask                                                          \\ \hline
Thank you.              & Thank you., Thanks.                                                  \\ \hline
neighbour (BE)          & neighbour, neighbor                                                  \\ \hline
... ¿no?                & ¿no, ¿verdad                                                         \\ \hline
escuchar algo           & escuchar algo, escuchar, escuchar (algo)                             \\ \hline
todo/-a                 & todo/-a, todo, toda, todo toda, toda todo, todoa                     \\ \hline
soy                     & soy, yo soy, soy estoy, estoy soy, yo soy estoy, yo estoy soy        \\ \hline
la télé / la télévision & la télé, la télévision, la télé la télévision, la télévision la télé \\ \hline
l'effaceur (m.)         & l'effaceur, l'effaceur m                                             \\ \hline
toi                     & toi, tu                                                              \\ \hline
gratuit/gratuite        & gratuit, gratuite, gratuit gratuite, gratuite gratuit                \\ \hline
\end{tabular}}
\caption{Examples of the list of correct possibilities, which can be compared with user answers to find possible matches.}
\vspace*{-3mm}

\label{tab:possibility_examples}
\end{table}

\subsubsection{Edit Score to Identify Synonyms}
The drawback of using such possibility lists as described above is missing synonyms and other correct solutions in the user data and mislabeling them as Wrong. The edit distance is used to filter for such possible cases and then inspect and label this data by hand. The edit distance is a valuable tool, as most incorrect solutions include typos and wrong letters and are thus, by edit distance, very close to the correct set of solutions. We choose the Levenshtein distance~\cite{levenshtein1966binary} as edit distance.

By computing the edit distance and filtering words with a high edit ratio, it is possible to identify significantly different words to the correct solution. These words are likely to be synonyms. A list of possible synonyms and correct solutions could be identified and then labeled by hand. Examples can be found in Table~\ref{tab:synoym_examples}.

\begin{table}[h]
\centering
\resizebox{0.45\textwidth}{!}{%
\begin{tabular}{|l|l|l|}
\hline
\textbf{Question} & \textbf{Answer} & \textbf{Synonym / Correct Solution} \\ \hline
um halb acht      & at 7:30         & at half past seven                  \\ \hline
richtig; korrekt  & right           & correct                             \\ \hline
die Pop-Musik     & el pop          & la música pop                       \\ \hline
Wie geht´s?       & ¿qué tal?       & ¿cómo estás?                        \\ \hline
braun             & brun/brune      & marron marron                       \\ \hline
ein Stadtviertel  & un quartier     & un arrondissement                   \\ \hline
\end{tabular}}
\caption{Examples of synonyms and additional correct answers detected with edit distance.}
\label{tab:synoym_examples}
\end{table}

\section{Data Set Format} \label{sec:format}
\begin{figure}[h]
\centering
\includegraphics[width=0.48\textwidth]{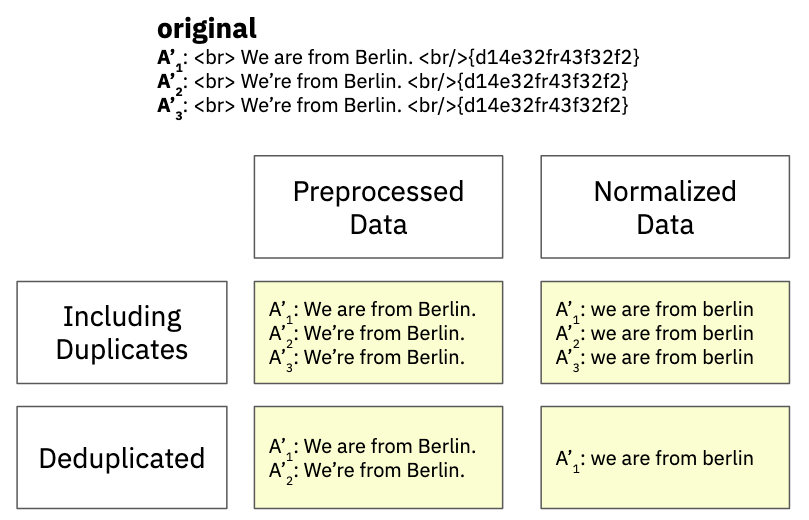}
\caption{Data set variations with example.}
\label{fig:variations}
\end{figure}

There are different preprocessed variations of the data set. The data is either preprocessed (remove HTML tags and sound IDs present in the export) or normalized as described in Section~\ref{sec:norm}. We present preprocessed data because it contains the format of possible user answers and generalizes more. In contrast, when using normalized data, the data is maximally ``clean''. Another aspect of the data set format is the inclusion of duplicates. Keeping duplicates might highlight the importance of common mistakes, while it could also lead to a system only adapting to these common mistakes and not generalizing enough. We decided to generate four different variations of the data set to determine the best resulting machine learning model experimentally. An example of these variations is visualized in Figure~\ref{fig:variations}.

\subsection{Balancing Data Set}
The data set variations including duplicates are balanced sufficiently (see Section \ref{sec:analysis}). On the other hand, the deduplicated data set variations are extremely imbalanced. This distribution is to be expected; given that, for one vocabulary question, there are many more possible incorrect answers than correct solutions. Thus, a far higher percentage of the deduplicated user answers is \textbf{False}, and there are only a few correct variations for the \textbf{True} class. The deduplicated data sets are undersampled to achieve fully balanced data sets.

\subsection{Disjoint Test Set \label{sec:testset}}

The most learned 1,001 to 1,250 vocabulary cards and their user answers are used to create an additional disjoint test set. The data is processed and labeled the same way as the training and validation data set. We again create the same four variations (preprocessed vs. normalized and duplicated vs. deduplicated). Furthermore, the deduplicated variations are undersampled. Models are to be evaluated on these test sets to gain more insights into their performance on new vocabulary cards closer to the real-life scenario.

\subsection{Availability}
The data set is available on European Language Grid\footnote{\url{https://live.european-language-grid.eu/catalogue/corpus/9487}}. It is split up in four tab separated files, one for each variation, per train and test set. The files include the following columns:
\begin{itemize}
    \item \textbf{cardId} - numeric card ID
    \item \textbf{question} - vocabulary card question
    \item \textbf{answer} - vocabulary card answer
    \item \textbf{userAnswer} - user answer input
    \item \textbf{Label} - True if user answer is correct, False if it is not
    \item \textbf{language} - target language (English, French or Spanish)
\end{itemize}

The processed data set variations include the following additional columns:
\begin{itemize}
    \item \textbf{question\_norm} - question normalized
    \item \textbf{answer\_norm} - answer normalized
    \item \textbf{userAnswer\_norm} - user answer normalized
\end{itemize}

The  deduplicated processed data sets do not include the \textbf{userAnswer}, since we deduplicate on basis of the \mbox{\textbf{userAnswer\_norm}} column. 

\section{Data Analysis} \label{sec:analysis}

\begin{table*}[t!]
\centering
\resizebox{\textwidth}{!}{%
\begin{tabular}{|l|c|c|c|c|c|c|c|c|}
\hline
\multirow{2}{*}{}        & \multicolumn{2}{c|}{\textbf{Preprocessed + Duplicates}}                                                                    & \multicolumn{2}{c|}{\textbf{Preprocessed + No Duplicates}}                                                           & \multicolumn{2}{c|}{\textbf{Normalized + Duplicates}}                                                                & \multicolumn{2}{c|}{\textbf{Normalized + No Duplicates}}                                                     \\ \cline{2-9} 
                         & \textbf{True}                                            & \textbf{False}                                           & \textbf{True}                                         & \textbf{False}                                        & \textbf{True}                                            & \textbf{False}                                           & \textbf{True}                                        & \textbf{False}                                       \\ \hline
\multirow{2}{*}{English} & \multicolumn{2}{c|}{12,718,244}                                                                                     & \multicolumn{2}{c|}{89,268}                                                                                   & \multicolumn{2}{c|}{12,718,244}                                                                                     & \multicolumn{2}{c|}{3,310}                                                                                  \\ \cline{2-9} 
                         & \begin{tabular}[c]{@{}c@{}}6,186,558\\ 49\%\end{tabular} & \begin{tabular}[c]{@{}c@{}}6,531,686\\ 51\%\end{tabular} & \begin{tabular}[c]{@{}c@{}}44,634\\ 50\%\end{tabular} & \begin{tabular}[c]{@{}c@{}}44,634\\ 50\%\end{tabular} & \begin{tabular}[c]{@{}c@{}}6,186,558\\ 49\%\end{tabular} & \begin{tabular}[c]{@{}c@{}}6,531,686\\ 51\%\end{tabular} & \begin{tabular}[c]{@{}c@{}}1,655\\ 50\%\end{tabular} & \begin{tabular}[c]{@{}c@{}}1,655\\ 50\%\end{tabular} \\ \hline
\multirow{2}{*}{French}  & \multicolumn{2}{c|}{8,027,831}                                                                                      & \multicolumn{2}{c|}{59,538}                                                                                   & \multicolumn{2}{c|}{8,027,831}                                                                                      & \multicolumn{2}{c|}{4,078}                                                                                  \\ \cline{2-9} 
                         & \begin{tabular}[c]{@{}c@{}}2,972,047\\ 37\%\end{tabular} & \begin{tabular}[c]{@{}c@{}}5,055,784\\ 63\%\end{tabular} & \begin{tabular}[c]{@{}c@{}}29,769\\ 50\%\end{tabular} & \begin{tabular}[c]{@{}c@{}}29,769\\ 50\%\end{tabular} & \begin{tabular}[c]{@{}c@{}}2,972,047\\ 37\%\end{tabular} & \begin{tabular}[c]{@{}c@{}}5,055,784\\ 63\%\end{tabular} & \begin{tabular}[c]{@{}c@{}}2,039\\ 50\%\end{tabular} & \begin{tabular}[c]{@{}c@{}}2,039\\ 50\%\end{tabular} \\ \hline
\multirow{2}{*}{Spanish} & \multicolumn{2}{c|}{2,248,457}                                                                                      & \multicolumn{2}{c|}{31,838}                                                                                   & \multicolumn{2}{c|}{2,248,457}                                                                                      & \multicolumn{2}{c|}{3,858}                                                                                  \\ \cline{2-9} 
                         & \begin{tabular}[c]{@{}c@{}}811,918\\ 36\%\end{tabular}   & \begin{tabular}[c]{@{}c@{}}1,436,539\\ 64\%\end{tabular} & \begin{tabular}[c]{@{}c@{}}15,919\\ 50\%\end{tabular} & \begin{tabular}[c]{@{}c@{}}15,919\\ 50\%\end{tabular} & \begin{tabular}[c]{@{}c@{}}811,918\\ 36\%\end{tabular}   & \begin{tabular}[c]{@{}c@{}}1,436,539\\ 64\%\end{tabular} & \begin{tabular}[c]{@{}c@{}}1,929\\ 50\%\end{tabular} & \begin{tabular}[c]{@{}c@{}}1,929\\ 50\%\end{tabular} \\ \hline
\end{tabular}%
}
\caption{Distribution of data points per language and data set variation for the training set.}
\label{tab:train_val_stat}
\end{table*}

\begin{table*}[t!]
\centering
\resizebox{\textwidth}{!}{%
\begin{tabular}{|l|c|c|c|c|c|c|c|c|}
\hline
\multirow{2}{*}{}        & \multicolumn{2}{c|}{\textbf{Preprocessed + Duplicates}}                                                                    & \multicolumn{2}{c|}{\textbf{Preprocessed + No Duplicates}}                                                         & \multicolumn{2}{c|}{\textbf{Normalized + Duplicates}}                                                                & \multicolumn{2}{c|}{\textbf{Normalized + No Duplicates}}                                                 \\ \cline{2-9} 
                         & \textbf{True}                                            & \textbf{False}                                           & \textbf{True}                                        & \textbf{False}                                       & \textbf{True}                                            & \textbf{False}                                           & \textbf{True}                                      & \textbf{False}                                     \\ \hline
\multirow{2}{*}{English} & \multicolumn{2}{c|}{2,329,762}                                                                                      & \multicolumn{2}{c|}{15,260}                                                                                 & \multicolumn{2}{c|}{2,329,762}                                                                                      & \multicolumn{2}{c|}{882}                                                                                \\ \cline{2-9} 
                         & \begin{tabular}[c]{@{}c@{}}1,155,401\\ 50\%\end{tabular} & \begin{tabular}[c]{@{}c@{}}1,146,584\\ 50\%\end{tabular} & \begin{tabular}[c]{@{}c@{}}7,630\\ 50\%\end{tabular} & \begin{tabular}[c]{@{}c@{}}7,630\\ 50\%\end{tabular} & \begin{tabular}[c]{@{}c@{}}1,155,401\\ 50\%\end{tabular} & \begin{tabular}[c]{@{}c@{}}1,146,584\\ 50\%\end{tabular} & \begin{tabular}[c]{@{}c@{}}441\\ 50\%\end{tabular} & \begin{tabular}[c]{@{}c@{}}441\\ 50\%\end{tabular} \\ \hline
\multirow{2}{*}{French}  & \multicolumn{2}{c|}{1,243,814}                                                                                      & \multicolumn{2}{c|}{9,478}                                                                                  & \multicolumn{2}{c|}{1,243,814}                                                                                      & \multicolumn{2}{c|}{944}                                                                                \\ \cline{2-9} 
                         & \begin{tabular}[c]{@{}c@{}}484,651\\ 39\%\end{tabular}   & \begin{tabular}[c]{@{}c@{}}759,163\\ 61\%\end{tabular}   & \begin{tabular}[c]{@{}c@{}}4,739\\ 50\%\end{tabular} & \begin{tabular}[c]{@{}c@{}}4,739\\ 50\%\end{tabular} & \begin{tabular}[c]{@{}c@{}}484,651\\ 39\%\end{tabular}   & \begin{tabular}[c]{@{}c@{}}759,163\\ 61\%\end{tabular}   & \begin{tabular}[c]{@{}c@{}}472\\ 50\%\end{tabular} & \begin{tabular}[c]{@{}c@{}}472\\ 50\%\end{tabular} \\ \hline
\multirow{2}{*}{Spanish} & \multicolumn{2}{c|}{388,923}                                                                                        & \multicolumn{2}{c|}{6308}                                                                                   & \multicolumn{2}{c|}{388,923}                                                                                        & \multicolumn{2}{c|}{1042}                                                                               \\ \cline{2-9} 
                         & \begin{tabular}[c]{@{}c@{}}388,923\\ 41\%\end{tabular}   & \begin{tabular}[c]{@{}c@{}}228,326\\ 59\%\end{tabular}   & \begin{tabular}[c]{@{}c@{}}3,154\\ 50\%\end{tabular} & \begin{tabular}[c]{@{}c@{}}3,154\\ 50\%\end{tabular} & \begin{tabular}[c]{@{}c@{}}388,923\\ 41\%\end{tabular}   & \begin{tabular}[c]{@{}c@{}}228,326\\ 59\%\end{tabular}   & \begin{tabular}[c]{@{}c@{}}521\\ 50\%\end{tabular} & \begin{tabular}[c]{@{}c@{}}521\\ 50\%\end{tabular} \\ \hline
\end{tabular}%
}
\caption{Distribution of data points per language and data set variation for the test set.}
\label{tab:test_stat}
\end{table*}

Table \ref{tab:train_val_stat} shows the number of user answers for the top 1,000 most learned vocabulary cards per language and variation in the final data set. The total number of data points and the number of user answers per class are shown. 

The data set variations, including duplicates, contain the most data points, as to be expected. English is the largest of the three languages as English is the most popular foreign language in the German school system, followed by French and Spanish. The duplicated data set variations are relatively balanced. Deduplicated data set variations were undersampled to achieve balanced data sets. The data sets are significantly smaller due to the undersampling of the larger possible set of False answers. Normalizing the user answers decreases the size of the data set even further since the set of possible correct answers is smaller.

The same insights can be deducted for the test set (Table \ref{tab:test_stat}). The test is smaller as it contains only the user answers to 250 vocabulary cards. 

\section{Experiments and Validation} \label{sec:exp}
We conduct experiments to explore the task of vocabulary evaluation and establish the usability of the data set. Eventually, we fine-tune a pre-trained BERT (Bidirectional Encoder Representations from Transformers)~\cite{devlin2018bert} model using the described data set as a downstream task. We compare the results for the data set variations and individual languages. 

\subsection{Parameters}
We fine-tune BERT models pre-trained for each language to ensure compatibility with each language. For English, we use Vanilla BERT$_{BASE}$~\cite{devlin2018bert}, for French CamemBERT~\cite{martin2019camembert}, and for Spanish BETO$_{BASE}$~\cite{canete2020spanish}. In addition, we also fine-tune a multilingual BERT (mBERT)~\cite{devlin2018bert} model with the concatenation of all languages.

In terms of parameters for fine-tuning the pre-trained models, we experimented with the hyperparameter space suggested by~\newcite{devlin2018bert}. Eventually, the models were trained for 4 epochs, using a batch size of 32 (16 for the English model). We used a learning rate of $3e-5$ for the English and Spanish model and $2e-5$ for the French and multilingual model. 

The data sets including duplicate user answers include up to more than 12 million data points (see Table~\ref{tab:train_val_stat}). Training a model with this amount of data leads to very long training times and the overfitting of the model to the training data. We downsampled the data for training and validation to 1 million data points to overcome this challenge while keeping distribution and structure in place. 

\subsection{Results}

\begin{figure}[H]
\centering
\includegraphics[width=0.48\textwidth]{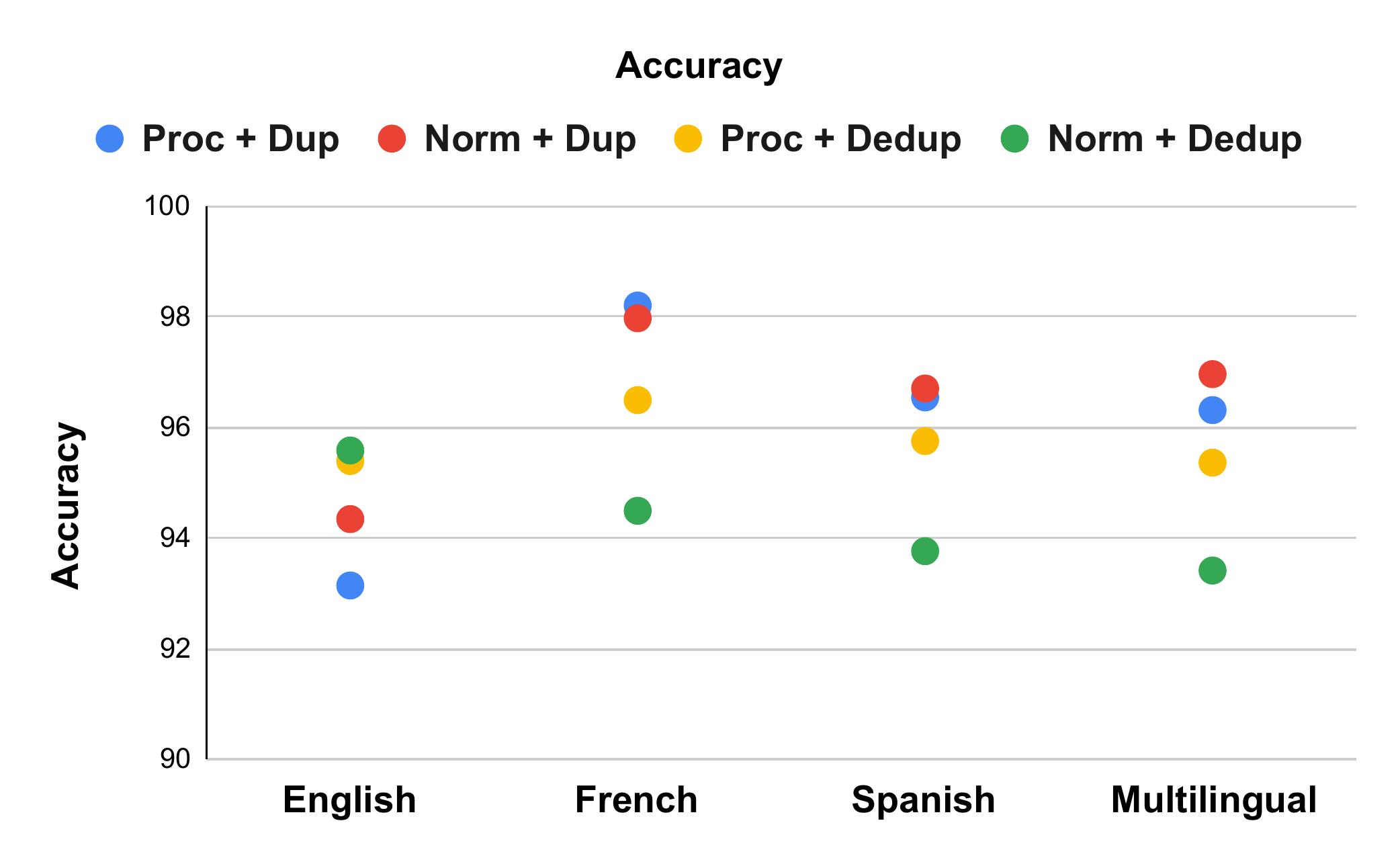}
\caption{Accuracy results: each data set reaches $> 92$ accuracy, and for each language, there also exists a model with $> 95.5$ accuracy, showing the models are able to learn from the available data.}
\label{fig:accuracy}
\end{figure}

\begin{figure}[H]
\centering
\includegraphics[width=0.48\textwidth]{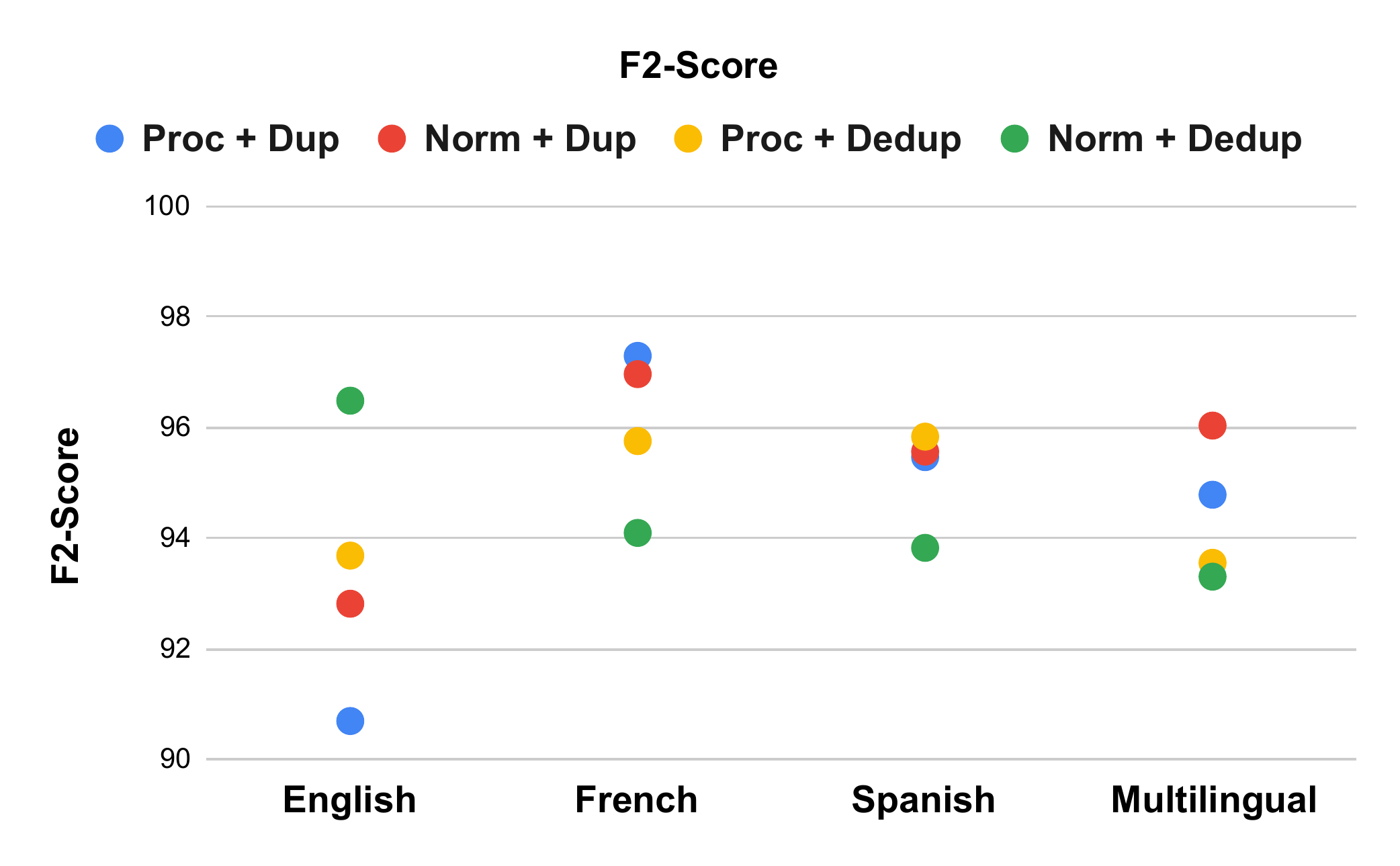}
\caption{F2-Score results: performance is comparable to accuracy, showing the models are able to balance precision and recall while highlighting recall.}
\label{fig:f2}
\end{figure}

The results from BERT fine-tuning for each language and data set variation are visualized in Figure \ref{fig:accuracy} and \ref{fig:f2}. We determined accuracy and F2-score to be the most relevant metrics. Accuracy indicates the overall quality of the model. F2-score, a variation of the F1-score, emphasizes the completeness (recall) of a system, which is important in vocabulary evaluation.

Overall, we can conclude that the fine-tuning results indicate excellent performance. For each data set variation in each language, we were able to fine-tune a model that reaches an accuracy of $> 92$. Further, a model with $> 95.5$ accuracy also exists for each language. These results show that the model learns to classify most vocabulary cards correctly. The high F2-scores confirm this finding. It shows that the model can learn from the available training data set and further generalize to classify new vocabulary cards correctly, demonstrated by the disjoint test set.

Most models could learn best from the data sets that include duplicates. Only Vanilla BERT seems to generalize better from deduplicated data. There is no clear performance distinction between the preprocessed and normalized data, which indicates that BERT can abstract from the textual input to perform the classification task. 

\section{Conclusion}
In this paper, we present a data set for the task of vocabulary evaluation called MuLVE. By using real-life pupil vocabulary training data, we are able to provide a data set for English, French, and Spanish in four variations. The primarily automated re-labeling process allows generating improved labels compared to existing language learning evaluation systems. A first experiment, fine-tuning pre-trained BERT models to the downstream task of vocabulary evaluation, shows excellent results. 

We aim to extend the data set to more languages in future work. Further, we aim to incorporate qualitative vocabulary evaluations to provide fine-grained feedback to language learners.

\section{Bibliographical References}\label{reference}
\label{main:ref}

\bibliographystyle{lrec2022-bib}
\bibliography{MuLVE}

\end{document}